\documentclass[10pt,twocolumn,letterpaper]{article}

\usepackage{cvpr}              

\usepackage{graphicx}
\usepackage{amsmath}
\usepackage{amssymb}
\usepackage{multirow}
\usepackage{booktabs}
\usepackage{color}
\usepackage[american]{babel}
\usepackage{microtype}

\usepackage{pifont}
\newcommand{\cmark}{\ding{51}}
\newcommand{\xmark}{\ding{55}}

\usepackage[pagebackref,breaklinks,colorlinks]{hyperref}

\newcommand{\squishlist}{
	\begin{list}{$\bullet$}
		{ \setlength{\itemsep}{0pt}
			\setlength{\parsep}{1pt}
			\setlength{\topsep}{1pt}
			\setlength{\partopsep}{0pt}
			\setlength{\leftmargin}{1.5em}
			\setlength{\labelwidth}{1em}
			\setlength{\labelsep}{0.5em} } }
\newcommand{\squishend}{\end{list} 
}

\usepackage[capitalize]{cleveref}
\crefname{section}{Sec.}{Secs.}
\Crefname{section}{Section}{Sections}
\Crefname{table}{Table}{Tables}
\crefname{table}{Tab.}{Tabs.}
\usepackage{color}

\def\cvprPaperID{****} 
\def\confName{CVPR}
\def\confYear{2022}

\begin{document}

\title{Relieving Long-tailed Instance Segmentation via Pairwise Class Balance}

\author{
	Yin-Yin He$^1$\footnotemark[1], 
	Peizhen Zhang$^2$\footnotemark[1] \footnotemark[2], 
	Xiu-Shen Wei$^{3,1}$, 
	Xiangyu Zhang$^{2}$, 
	Jian Sun$^{2}$ \\\\
$^1$State Key Laboratory for Novel Software Technology, Nanjing University \\
$^2$MEGVII Technology \\
$^3$School of Computer Science and Engineering, Nanjing University of Science and Technology \\
{\tt\small heyy@lamda.nju.edu.cn, weixs.gm@gmail.com} \\
{\tt\small \{zhangpeizhen, zhangxiangyu, sunjian\}@megvii.com}
}

\maketitle

\renewcommand{\thefootnote}{\fnsymbol{footnote}}

\footnotetext[1]{Equal contribution. This paper is supported by the National Key R\&D Plan of the Ministry of Science and Technology (Project No. 2020AAA0104400), CAAI-Huawei MindSpore Open Fund, and Beijing Academy of Artificial Intelligence (BAAI). This work is done during Yin-Yin He’s internship at MEGVII Technology.}
\footnotetext[2]{Corresponding author.}

\begin{abstract}
   Long-tailed instance segmentation is a challenging task due to the extreme imbalance of training samples among classes. It causes {severe biases of the head classes (with majority samples) against the tailed ones}. {This renders ``how to appropriately define and alleviate the bias'' one of the most important issues}. Prior works mainly use label distribution or mean score information to indicate a coarse-grained bias. In this paper, we {explore to excavate the confusion matrix}, which carries the fine-grained misclassification details, to relieve the pairwise biases, {generalizing the coarse one}. {To this end, we propose a novel Pairwise Class Balance (PCB) method, built upon a confusion matrix} which is updated during training to {accumulate} the ongoing {prediction preferences}. PCB generates fightback soft labels for regularization during training. Besides, an iterative learning paradigm is developed to {support a progressive and smooth regularization in such debiasing.} PCB can be plugged and played to any existing method as a complement. Experimental results on LVIS demonstrate that our method achieves state-of-the-art performance without bells and whistles. Superior results across various architectures show the {generalization ability}. The code and trained models are available at \url{https://github.com/megvii-research/PCB}.
\end{abstract}

\section{Introduction}
\label{sec:intro}

\begin{figure}
    \centering
    \includegraphics[width=1.0\columnwidth]{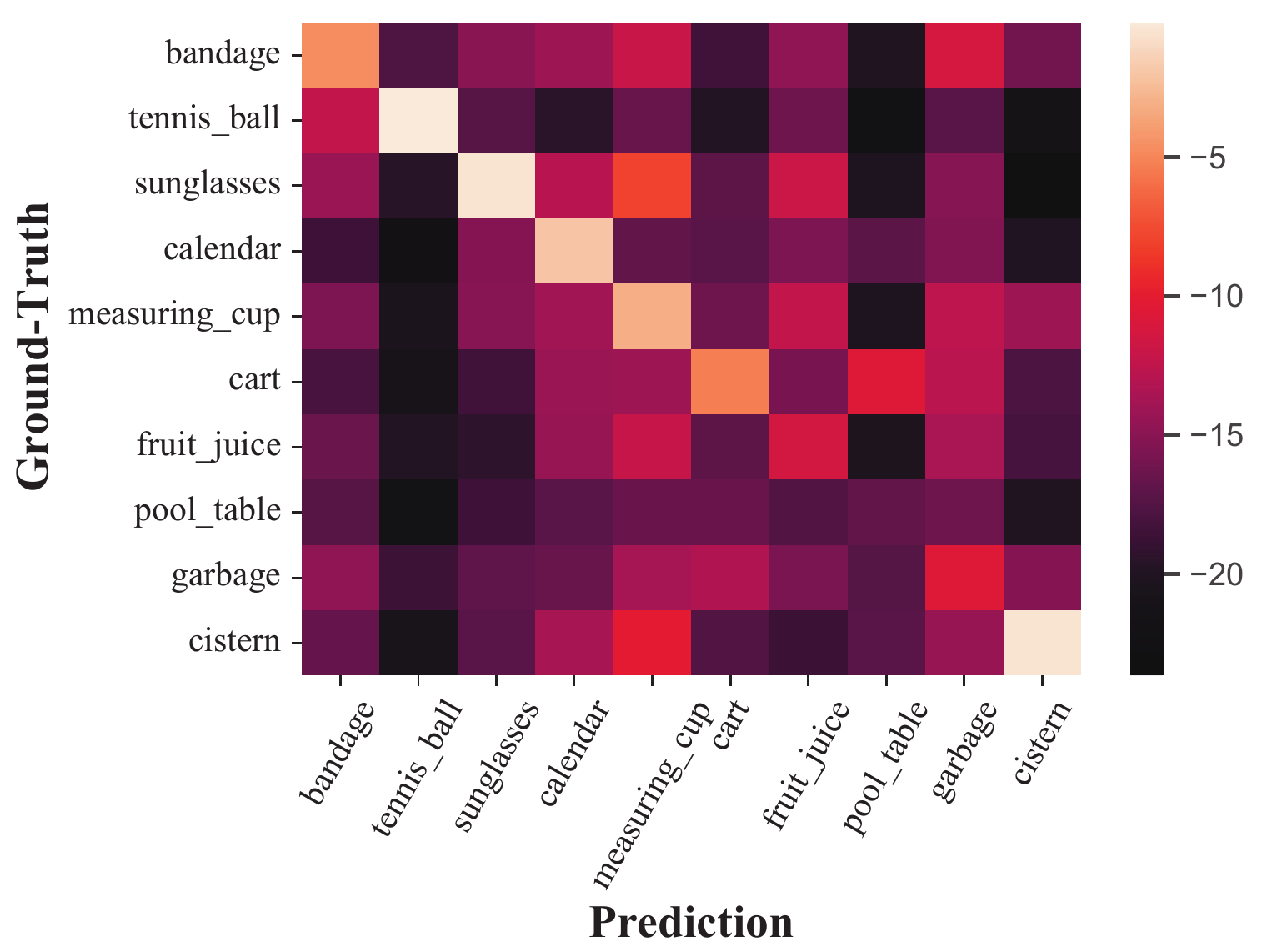}
    \caption{Visualization of a randomly sampled 10-class categorical confusion matrix of Mask R-CNN with ResNet-50-FPN on LVIS v0.5 after {taking logarithm with base} 2. {In current palette, higher probabilities are shown with lighter colors.} It reflects the model biases clearly (\textit{e.g.}, the probability of misclassifying \textit{sunglasses} to \textit{measuring cup} is larger than {the reverse}).} 
    
    \label{fig:intro_cm}
\end{figure}

The success of modern object detectors and instance segmentors has been verified on rich and balanced datasets. {However, their performance drops dramatically when applied to datasets like LVIS\,\cite{gupta2019lvis} which is closer to the long-tailed and large-vocabulary category distribution in the real-world scenario.} {The devil lies in the classification prediction bias caused by extreme imbalanced training sample volumes} among foreground classes \cite{wang2020devil}. {Typically, it is common to leverage instructive indications with positive correlation with the prediction bias into modeling for de-biasing and thus achieve better results.} {Prior works exploit the class-wise sample frequencies in the training set as an intuitive indication} \cite{gupta2019lvis, chang2021image, ren2020balanced, wang2021seesaw}. However, the learning quality of a class is not {only about the} distribution prior {but also factors like optimization hardness \cite{duggal2020elf}, relevant to model learning process.} {Feng \textit{et al}. proposed the} \textit{mean classification score} \cite{feng2021exploring} metric. Such {train-time} model statistics can reflect the learning quality of classes {beyond mere} label distribution. {Yet, we notice that it considers only the sample classification statistics within each class, ignoring the inter-class similarities.}

{To involve the inner-and-inter class relationship, the categorical} confusion matrix is already a weapon in hand. It carries the {dynamic misclassification conditional probability distribution} between pairs of classes (Fig.~\ref{fig:intro_cm}). A {proof study} is conducted on LVIS v0.5 \cite{gupta2019lvis} to verify our conjecture. The upper bound performance of post-hoc calibration using confusion matrix and mean score (both of the confusion matrix and mean score are collected on the validation set) is testified. Detailed results are summarized in Tab.~\ref{tab:motivation_tab}. {As shown}, calibration using mean score and confusion matrix can both lift the performance, especially for the rare classes. {More importantly}, the upper bound of confusion matrix calibration is much higher than that of mean score calibration (+ $\mathbf{9.4}$ {AP$_r$} on the random sampler and + $\mathbf{8.0}$ {AP$_r$} on RFS~\cite{gupta2019lvis} sampler with a similar performance on frequent classes). Confusion matrix calibration achieves almost fully unbiased performance. Please refer to Sec. \ref{subsec:softlabel} for details.

\begin{table}
   \caption{The performance of models before and after post-hoc calibration using different model statistics on the validation set. Experiments were conducted on LVIS v0.5 using Mask R-CNN with ResNet-50-FPN. MS stands for mean classification score \cite{feng2021exploring} and CM stands for confusion matrix. Calibration using CM could achieve almost fully unbiased performance.}
   \centering
   \small
   \begin{tabular}{cc|ccccc}
      \hline
	    Sampler & Info. & AP & AP$_r$ & AP$_c$ & AP$_f$ & AP$^b$ \\
	   \hline \hline
	   \multirow{3}{*}{Random} & / & 21.9 & 4.6 & 21.7 & \textbf{28.9} & 21.9 \\
	   & MS & 25.7 & 14.3 & \textbf{27.3} & 28.2 & 25.4 \\
	   & CM & \textbf{26.3} & \textbf{23.7} & 26.0 & 27.9 & \textbf{25.9} \\
	   \hline
	   \multirow{3}{*}{RFS \cite{gupta2019lvis}} & / & 25.6 & 16.0 & 26.4 & \textbf{28.5} & 25.6 \\
	   & MS & 27.0 & 20.6 & 28.2 & 28.0 & 27.0 \\
	   & CM & \textbf{28.4} & \textbf{28.6} & \textbf{28.8} & 27.9 & \textbf{28.2} \\
	   \hline 
   \end{tabular}
   \label{tab:motivation_tab}
\end{table}

Accordingly, the fine-grained misclassification probabilities between pairwise classes in the confusion matrix, {which we conclude as} \textit{pairwise bias}, {is powerful as} an indicator. One vital issue which has to be resolved is that the validation set is unavailable in practical training. One straightforward intuition is to utilize instead the confusion matrix over the train set for calibration. Unfortunately, this {malfunctions} as shown in Tab.~\ref{tab:component_ablation}. This is probably due to the mismatch between the confusion matrix calculated at train and test time, which is originated from {diversified patterns of samples}.

Inspired by \cite{hong2021disentangling} who conducted train-time disentangling to replace test-time post compensation for long-tailed classification, we develop an online {pairwise bias-driven} calibration method named PCB (\textbf{P}airwise \textbf{C}lass \textbf{B}alance). {It maintains a confusion matrix during training with fightback targets generated in a matrix-transposed posterior manner to balance the pairwise bias for each ongoing proposal learning.} However, {naive exploitation might limit the efficacy}, as stronger regularization {could be detrimental to the} discrimination {ability} while weaker regularization can not relieve the pairwise bias well. To fully absorb the merits of the above PCB regularization and {meanwhile facilitate the debiasing}, a light prediction-dependent iterative  paradigm is equipped. {It is accomplished by using the discriminative predictions trained by the original one-hot labels to be embedded back to enhance the features recurrently. PCB regularization is gradually applied to the predictions from each step of the enhanced feature.} In this way, {a more friendly, progressive pairwise class balancing is achieved} which is also proven to be effective in later experimental sections. 

To summarize, our contributions are as follows:

\squishlist
    \item We explore the use of confusion matrix to indicate pairwise model bias in the field of long-tailed instance segmentation, which shows a promising upper bound.
    \item An Pairwise Class Balance method is proposed to tackle the long-tailed instance segmentation.
    \item Extensive experiments on LVIS v0.5 and LVIS v1.0 show the effectiveness of our method. 
\squishend

\section{Related Work}

\textbf{Object Detection and Instance Segmentation.}
Object detection has attracted lots of attention in recent years, with remarkable improvements being made. Modern object detection frameworks \cite{girshick2015fast, ren2015faster, cai2018cascade, lin2017focal, tian2019fcos, zhang2020bridging} can be divided into two-stage and one-stage ones. The two-stage detectors \cite{girshick2015fast, ren2015faster, cai2018cascade} first generate a set of proposals in the first stage, then refine the proposals and perform classification in the second stage. While one-stage detectors \cite{lin2017focal, tian2019fcos, zhang2020bridging} directly predict bounding boxes. Compared to two-stage detectors, one-stage detectors are faster, yet two-stage detectors can provide better localization. 

Mask R-CNN \cite{he2017mask} adapts Faster R-CNN \cite{ren2015faster} to instance segmentation task by adding a mask prediction branch in the second stage. SOLO \cite{wang2020solo, wang2020solov2} is another line of instance segmentation that is box-free. Our work is based on Mask R-CNN to stay the same as other long-tailed instance segmentation works.

\textbf{Long-tail Learning.}
Long-tailed learning problem is across various domains (e.g., fine-grained recognition \cite{van2018inaturalist}, multi-label learning \cite{wu2020distribution} and instance segmentation \cite{gupta2019lvis}). Two classic solutions are data re-sampling \cite{chawla2002smote, gupta2019lvis} which aims to flatten the data distribution and loss re-weighting \cite{cui2019class, cao2019learning} which emphasizes more on tail data. Recent works proposed decoupling training \cite{kang2019decoupling, zhou2020bbn} which first obtains a good representation through conventional training and then calibrates the classifier. Other techniques like ensemble \cite{wang2020long}, self-supervised learning \cite{yang2020rethinking, li2021self} and knowledge distillation \cite{li2021self, he2021distilling} are verified to be useful in long-tailed learning.

\cite{gupta2019lvis} first introduced the long-tailed learning problem to instance segmentation and thus to object detection. They built a large vocabulary dataset named LVIS and proposed a simple baseline RFS. \cite{wang2020devil} pointed out that the long-tail property affects classification most. Later on, a series of works tried to alleviate classification bias. One line of works tried to improve the sample strategy \cite{chang2021image, wu2020forest, feng2021exploring, zhang2021mosaicos, zhang2021bag}, while another major line of works focus on loss engineering. Equalization Loss \cite{tan2020equalization} and its improvements \cite{tan2021equalization} down-weight the negative gradients for tail classes from head classes, while droploss \cite{hsieh2021droploss} further takes the gradients from background into consideration. Similarly, ACSL \cite{wang2021adaptive} only penalize negative classes over threshold. Separating the categories into some small groups \cite{wu2020forest, li2020overcoming} and simple calibration \cite{zhang2021distribution, pan2021model} helps, too. \cite{ren2020balanced, wang2021seesaw} modified the original soft-max function by embedding the distribution prior, achieving success. \cite{feng2021exploring} first introduce model statistics., it utilized mean classification scores in place of the model-agnostic prior. It fails to point out the direction the prediction of one class biased to. So we take one step further, a two-dimensional statistics (i.e., confusion matrix) is utilized to indicate fine-grained pairwise bias.

\textbf{Confusion Matrix.} Confusion matrix is a classical tool for error analysis. In many fields, it has shown powerful abilities. It's used to estimate the target distribution under label shift \cite{lipton2018detecting}. In the field of label noise, rather than hard prediction, \cite{patrini2017making} uses soft prediction on the cleanest sample of each class to generate a confusion matrix and to assume the noisy ratio. Similarly, \cite{zhang2021delving} actually keeps a confusion matrix using soft prediction for label smoothing. To the best of our knowledge, we are the first to adopt confusion matrix in the field of long-tailed learning.

\section{Methodology}

{
In this section, we first discuss the long-tailed phenomenon in a pairwise bias perspective revealed by the confusion matrix (cf. Sec. \ref{subsec:cm}). Next, a post-hoc calibration verification shows a promising upper bound of balancing the bias (cf. Sec. \ref{subsec:softlabel}). In realistic training, we propose an online iterative regularization paradigm to relieve the pairwise class bias that facilitates long-tailed instance segmentation (cf. Sec. \ref{subsec:ieterative}).}

\begin{figure*}[t]
    \centering
    \includegraphics[width=2.0\columnwidth]{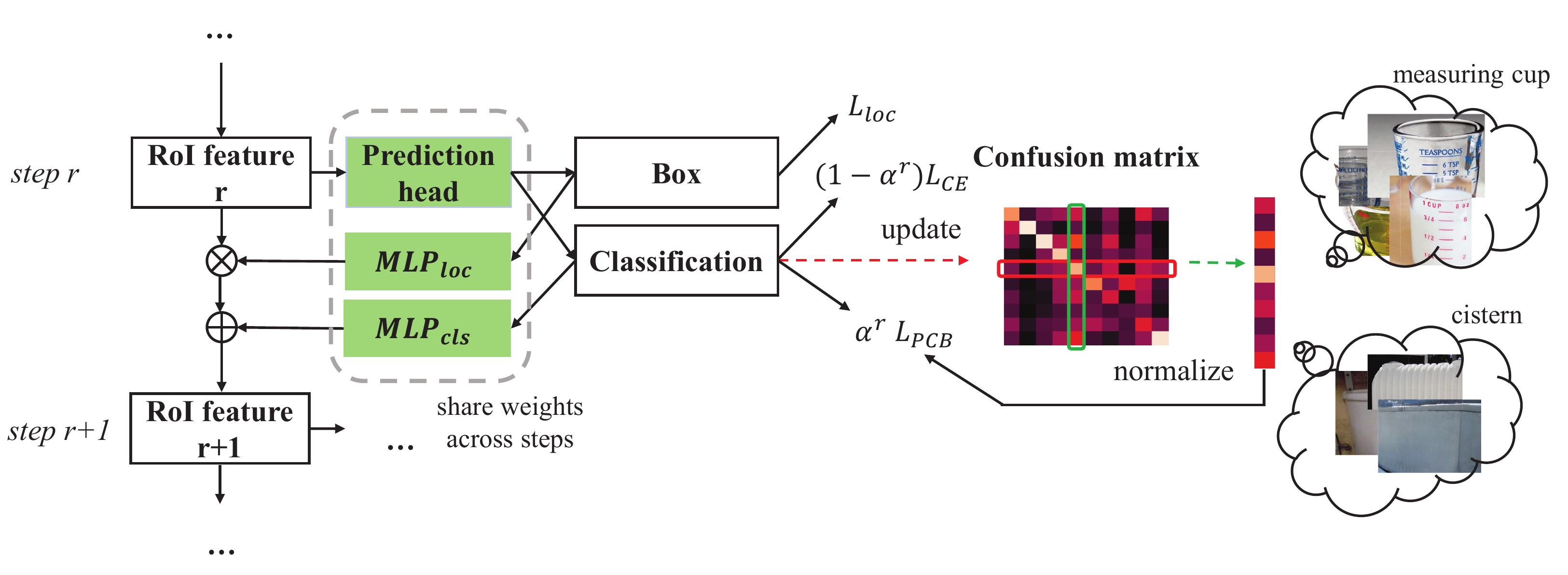}
    \caption{{Our proposed PCB framework: At each recurrent step $r$, the RoI features generated by the previous step are fed to the shared prediction head to obtain predicted boxes and scores. Soft targets are generated according to the confusion matrix for classification regularization $L_{PCB}$, traded-off by $\alpha_r$ over $L_{CE}$. Subsequently, the RoI features are updated by the predictions for next step usage, and the confusion matrix is updated by the current iteration of score statistics. }}
    \label{fig:framework}
\end{figure*}

\subsection{Confusion matrix on indicating pairwise bias}
\label{subsec:cm}

{{Most prior works dealing with long-tailed problems aim at relieving the prediction bias between data rich classes (i.e., head classes) and data scarce classes (i.e., tail classes).} They mainly convey the spirit of data re-sampling or loss re-weighting. However, these are confined to sample-level, without considering the model learning dynamics, and could be sub-optimal. LOCE \cite{feng2021exploring} proposed to use \textit{mean classification score} for each category across training to reflect the run-time predictive preference. Whereas, it failed to utilize the inter-class relationship that is critical in long-tailed literature. Instead, in this paper, we propose to leverage the classification confusion matrix to indicate the learning preference which we elaborate on in the following.}

{For ease of illustration and taking the classical two-stage instance segmentation model Mask R-CNN\,\cite{he2017mask} for instance, we denote by $F(\cdot)$ the R-CNN classification head. Without loss of generality, we mainly discuss the most-adopted cross-entropy loss (CE) (see also the applicability to binary cross-entropy in the last paragraph of Sec.~\ref{subsec:ieterative}). Normally, by taking as input a proposal feature map $X$, $F$ predicts a categorical distribution $\mathbf{z}=F(X) \in \mathbb{R}^{C+1}$ ($C$ foreground classes plus 1 background class). 
Since misclassification is accustomed to being among foreground classes under long-tailed setting\,\cite{tang2020long}, we investigate the foreground classes only. This is achieved by excluding the background logit ($\mathbf{z}^{fg} \in \mathbb{R}^{C}$) and re-normalizing the multinomial class probability below.}

\begin{equation}
    \hat{p}_i = \frac{\exp (z^{fg}_i)}{\sum_{k=1}^C \exp (z^{fg}_k)}\,.
\end{equation}

{
We denote $M\in \mathbb{R}^{C \times C}$ as the confusion matrix which could be calculated given histogram votes in two-dimensional bins of label-to-prediction statistics:}
\begin{equation}
\label{eq:cm_hard}
    M_{i,j} = \frac{\sum_{(\mathbf{x}, y)} \mathbb{I}[\arg\max(\mathbf{z}^{fg})={j,y}=i]}{\sum_{(\mathbf{x}, y)} \mathbb{I}[y=i]},
\end{equation}

{\noindent where $1\leq i,j \leq C$ and $\mathbb{I}[\cdot]$ serves as an indicator which evaluates 1 when the inner condition satisfies and 0 otherwise. To keep the finer misclassification distribution details (that are truncated by \textit{argmax} in Eq.~\ref{eq:cm_hard}) and also for more stable training, we opt for a softened version by aggregating the predicted probabilities at the ground-truth indice.}
\begin{equation}
    M_{i,j} = \frac{\sum_{(\mathbf{x}, y)} \hat{p}_{j} {\cdot}\mathbb{I}{[y=i]}}{\sum_{(\mathbf{x}, y)} \mathbb{I}{[y=i]}}\,.
    \label{eq:cm_soft}
\end{equation}

{
$M_{i,j}$ is the statistical probability, specifying to what extension a sample of class $i$ is classified as $j$ by the model. Unequal values between $M_{i, j}$ and $M_{j ,i}$ reflect the asymmetric model preference between the two classes. We term the phenomenon as \textit{pairwise bias}. The pairwise bias between class $i$ and $j$ becomes balanced when $M_{i, j} = M_{j ,i}$. Usually, in the long-tailed scenario, pairwise biases imbalance are more likely to happen between head and tail classes, with $M_{i,j} > M_{j,i}, \forall  i\in \mathcal{T}, j\in \mathcal{H}$ where sets of tail classes and head classes are denoted as $\mathcal{T}$ and $\mathcal{H}$ respectively. Under extreme cases, there could be $M_{i,j} \gg M_{j,i}, \exists i\in \mathcal{T}, j\in \mathcal{H}$. Officially, the head and tail classes are represented by three class splits, \textit{i.e.}, \textit{frequent} ($f$), \textit{common} ($c$) and \textit{rare} ($r$). Practically, for Mask R-CNN ResNet-50-FPN trained on LVIS v0.5, the probability of a frequent class instance being misclassified as a rare one, \textit{i.e.}, $M_{f, r}$ is 0.01 while the versus $M_{r, f}$ is 0.19.
}

\subsection{A post-hoc calibration trial}
\label{subsec:softlabel}

{Intuitive exploitation of the confusion matrix is for post-hoc calibration as mentioned in Sec. \ref{sec:intro}. Typically, a post-hoc calibration could be conducted by following the spirits of Bayes' total probability theorem.}

{
\begin{equation}
    P(y=i|x)=\sum_{j=1}^{C}{P(y=i|\hat{y}=j,x)P(\hat{y}=j| x)},
    \label{eq:total_prob_theorem}
\end{equation}

\noindent where condition $x$ omit $X=x$, representing a proposal feature. $y=i$ and $\hat{y}=j$ represent the events of ``$x$ belongs to class $i$ indeed'' and ``$x$ is predicted as class $j$'' respectively. Naturally, the $P(\hat{y}=j| x)$ term is instantiated by the predicted probability on $j$-th position ($\hat{p}_j$) of the classification output. We instantiate $P(y=i|\hat{y}=j,x)$ by $\hat{M}_{i,j}$ via performing normalization on the confusion matrix such that elements of each \textit{column} (not row) add up to 1.}
\begin{equation}
    {\hat{M}_{i, j}} = \frac{M_{i, j}}{\sum_{k=1}^C M_{k, j}}
    \label{eqa:cm_normalization}
\end{equation}

{$\hat{M}_{i,j}$ indicates how likely a sample is attributed to class $i$ overall, conditioned on being predicted as class $j$. It could be viewed as an approximate expectation of $P(y=i|\hat{y}=j,x)$. Hence, a post-hoc calibration of the predicted probability of sample $x$ upon class $i$, dubbed $\tilde{p}_i$, is corrected as:}

{
\begin{equation}
    \tilde{p}_i = \sum_{j=1}^C \hat{M}_{i, j} \hat{p}_j
    \label{eqa:cm_calibration}
\end{equation}}

{
In another view, the coefficient $\hat{M}_{i, j}$ together with $\hat{M}_{j, i}$ comprise a \textit{pairwise bias} defined in Sec.~\ref{subsec:cm}.

We also compare the above post-hoc calibration with the one aided by unary \textit{mean classification scores}\,\cite{feng2021exploring} below:
\begin{equation}
    s_i = \frac{\sum_{(\mathbf{x}, y)} \hat{p}_i \cdot\mathbb{I}[y = i]}{\sum_{(\mathbf{x}, y)} \mathbb{I}[y = i]}
\end{equation}
Apparently, they are exactly the diagonal elements in the confusion matrix (Eq.~\ref{eq:cm_soft}), \textit{i.e.}, $s_i = M_{i, i}$ (As a complementary, the extra pairwise bias items retain fine-grained inter-class misclassification hints, reflecting model preference in pairs of classes).}
{
Following the empirical findings \cite{feng2021exploring} that positive correlation exists between the mean classification score and the number of instances, we conduct below adjustment for calibration.
\begin{equation}
    \tilde{p}_{i} = \frac{\hat{p}_i/s_i}{\sum_{k=1}^C \hat{p}_k/s_k}
    \label{eq:mcs_calibration}
\end{equation}
}

{
Beyond above calibration via Eq.~\ref{eqa:cm_calibration} or ~\ref{eq:mcs_calibration} over foreground classes, we keep using the same background probability ($\tilde{p}_{C+1}=\hat{p}_{C+1}$) as suggested in \cite{tang2020long,wang2021seesaw}. To ensure the summation-1 property of the predictions, we first divide each foreground probability by their sum, rendering $\sum_{i=1}^{C}\tilde{p}_i=1$ (still using the same symbol with slight abuse of the notation). We then re-scale each foreground probability by multiplying coefficient $\delta=1-\tilde{p}_{C+1}$.
}

{We testify the post-hoc calibration efficacy \textit{w.r.t.} the two metrics (\textit{confusion matrix} or \textit{mean classification score}). 
Experiments are conducted upon LVIS v0.5 with confusion matrix computed on the validation set. See Tab.~\ref{tab:motivation_tab} for the details. 
For post-hoc evaluation which enables partial labels for upper bound examination, actual categories of the proposals are utilized to compute the confusion matrix (through matching using \textit{Intersection-over-Union} (IoU) between proposals and ground-truth boxes).
Notably, using confusion matrix for calibration is much more promising in respect of AP on rare classes ($\approx 10\%$ higher in random mode and similar to AP$_c$ and AP$_f$). Such advantages show a higher upper bound of confusion matrix against mean classification score by considering the non-diagonal \textit{pairwise biases}. This motivates our method in Sec.~\ref{subsec:ieterative} below.
}

\subsection{Online iterative confusion matrix learning}
\label{subsec:ieterative}

{The above post-hoc calibration relies on validation labels which are unavailable in usual training. Alternatively, one might use the confusion matrix statistics over the training set instead. However, the confusion matrix patterns on the training and validation set can not match perfectly, \textit{e.g.}, due to diversified object appearances even in the same category. Results in Tab.~\ref{tab:component_ablation} verify this experimentally. Inspired by \cite{hong2021disentangling} who conducted train-time disentangling to replace the test-time post compensation for classification, we propose to conduct online pairwise class balancing during training using the train-time confusion matrix information for long-tailed instance segmentation. See Fig.~\ref{fig:framework} for our framework.} 

{
We update the confusion matrix during training through exponential moving average (EMA). Specifically for the $t$-th minibatch, we update the matrix rows, each corresponding to a foreground class $y$ by the averaged prediction outputs of proposals assigned to it. 

\begin{equation}
    \mathbf{m}_{y}^t = \gamma \mathbf{m}_{y}^{t-1} + (1 - \gamma) \bar{\mathbf{p}}^{t}_{y}\,,
\end{equation}
where $\mathbf{m}_y^t$ and $\mathbf{m}_y^{t-1}$ denote the $y$-th row vector of the confusion matrix at $t$-1-th and $t$-th iteration, respectively. $\bar{\mathbf{p}}^{t}_{y}$ is the averaged predicted probability vector and $\gamma \in (0, 1)$ is the momentum. Updates to categories that do not appear in current iteration are skipped.
}

{
As specified in Sec.~\ref{subsec:softlabel}, the transformed confusion matrix items ($\hat{M}_{i,j}, 1\leq i,j\leq C$) by Eq.~\ref{eqa:cm_normalization} reflect the conditional posterior probability for rectification in the form of pairwise biases. We thus directly enforce the model to learn such information as a fightback regularization. Specifically, we leverage the transformed confusion matrix items as soft targets. 
Formally, for each $k$-th foreground proposal of label $y$ at current iteration, we apply a regularization upon foreground-classes learning via cross entropy below.}

{
\begin{equation}
    \label{eq:pcb_regularization}
    L_{PCB}(k) = - \sum_{i=1}^C \hat{M}_{i,y}^{t} \log \hat{p}_i
\end{equation}
}

{Beyond the rationale elaborated above, the regularization intuitively aims at balancing the pairwise biases. Taking a pair of classes $a$ and $b$ for illustration, the predicted confidence on class $a$ and $b$ is suppressed and raised simultaneously if there exists a preference $M_{a, b} < M_{b, a}$.}

{
The regularization aims at relieving train-time pairwise class balance in the macro model dynamics level. We keep the plain cross-entropy loss $L_{CE}(\cdot)$ for the micro sample level classification. Denote $K$ the number of proposals in current minibatch, the total classification loss function becomes $L_{cls}=\sum_{k=1}^{K}{L_{cls}(k)}$ with each term as:

\begin{equation}
    L_{cls}(k) = 
    \begin{cases}
        \alpha L_{PCB}(k) + (1 - \alpha) L_{CE}(k), & y_k \neq {C+1} \\
        L_{CE}(k), & otherwise \\
    \end{cases}\,,
    \label{eq:overall_cls}
\end{equation}
where $\alpha$ is a coefficient to trade-off the two loss functions.}

{
As shown in Tab.~\ref{tab:alpha_analysis}, raw exploitation of PCB regularization (Eq.~\ref{eq:pcb_regularization}) verifies the profits. Whereas, the performance deteriorates as the regularization term becomes dominant ($\alpha \geq 0.2$). We consider this is because current predictions are less discriminative at the first time to achieve the joint goal of the pairwise bias balancing and fundamental classification. Motivated by this and also the refinement inspiration in \cite{chu2020detection} (used to distinguish duplicated boxes in crowd detection), we propose a light prediction-dependent iterative paradigm to fully absorb the merits by PCB regularization. The core lies in using predictions stemming from a proposal feature to refine itself (then give a finer prediction next). The predicted boxes and confidences are projected as sparse spatial and channel attention to enhance the proposal feature, rather than direct concatenation as in \cite{chu2020detection} which could lead to semantic gaps. See the supplementary material for detailed analysis. Overall, Tab.~\ref{tab:component_ablation} show the magic complementary benefits. The whole process takes place only in the final head part and remains the proposal box unchanged, unlike \cite{cai2018cascade}. Technically, for each step $r$:}

{
\begin{enumerate}
    \item Calculating output classification logits $\mathbf{z}^r \in \mathbb{R}^{C+1}$ and predicted box $\mathbf{b}^r \in \mathbb{R}^{4C}$ with regard to classification and box head, given compact proposal feature map $X^r \in \mathbb{R}^{H_p \times W_p \times D}$ ($H_p$, $W_p$ and $D$ are pooled $7\times7$ resolution and channel depth). $X^0$ ($r=0$) is the regular proposal feature $X$.
    
    \item Computing the regular localization loss $L_{loc}^r$ together with the modified classification loss $L_{cls}^r$ (Eq.~\ref{eq:overall_cls}).
    
    \item Using separate MLPs (FC-ReLU-FC) to project $\mathbf{z}^r$ and $\mathbf{b}^r$ as D- and $H_pW_p$-dimensional features. Namely, obtaining $\mathbf{f}^{r}_{z}=MLP_{cls}(\mathbf{z}^r)$ and $\mathbf{f}^{r}_{b}=MLP_{loc}(\mathbf{b}^r)$ which are then warped into shapes of $\mathbb{R}^{1 \times 1 \times D}$ and $\mathbb{R}^{H \times W \times 1}$ respectively, denoted by $X^r_z$ and $X^r_b$. 
    
    \item Refining proposal feature $X^{r+1} = X_b^r \otimes X^r \oplus X_z^r$. $\otimes$ and $\oplus$ denote broadcast element-wise multiplication and addition, respectively.
\end{enumerate}
}

{Naturally, proposal features in later refinement steps become more discriminative and could bear stronger regularization. We simply impose a linear increase of $\alpha_r=\frac{r-1}{R-1}\alpha$ as steps move on. Additionally, an outer step-wise coefficient $w_r$ is introduced, incorporating both the classification and localization loss and we derive the overall objective:

\begin{equation}
    L = \left[\sum_{r=1}^R w_r (L_{cls}^r + L_{loc}^r)\right] + L_{mask}\,,
\end{equation}

The mask prediction loss is calculated once since the segmentation sub-task has its own branch.

\begin{table*}
   \caption{Results on the validation set of LVIS v0.5 and LVIS v1.0 with a ResNet-50 backbone. Our PCB can be complementary to various methods, including the baseline method Softmax, sampling methods (e.g., RFS \cite{gupta2019lvis}), one with binary classifier (e.g., EQL v2 \cite{tan2021equalization}), and even strong state-of-the-art methods (e.g., Seesaw \cite{wang2021seesaw}). The results of RFS and EQL v2 on LVIS v1.0 are direct copied from \cite{feng2021exploring}, and the results of Seesaw on LVIS v1.0 are directly copied from \cite{chen2019mmdetection}.}
   \centering
   \small
   \begin{tabular}{lc|cccccr|cccccr}
      \hline
        \multicolumn{2}{c|}{Dataset} & \multicolumn{6}{c|}{LVIS v0.5} & \multicolumn{6}{c}{LVIS v1.0} \\
	    Method & PCB & AP & AP$_r$ & AP$_c$ & AP$_f$ & AP$^b$ & \textit{PwB} $\downarrow$ & AP & AP$_r$ & AP$_c$ & AP$_f$ & AP$^b$& \textit{PwB} $\downarrow$\\
	   \hline \hline
	   \multirow{2}{*}{Softmax} & \xmark  & 21.9 & 4.6 & 21.7 & 28.9 & 21.9 & 13.8$\quad$ & 19.0 & 1.3 & 16.7 & 29.3 & 19.9 & 14.5$\quad$ \\
	   & \cmark & {\textbf{25.1}} & {\textbf{12.6}} & {\textbf{25.5}} & {\textbf{29.5}} & {\textbf{25.2}} & {\textbf{8.4}}$\quad$ & {\textbf{22.6}} & {\textbf{7.7}} & {\textbf{21.8}} & {\textbf{29.9}} & {\textbf{24.1}} & \textbf{8.8}$\quad$ \\
	   
	   \hline 
	   
	   \multirow{2}{*}{RFS \cite{gupta2019lvis}} & \xmark & 25.6 & 16.0 & 26.4 & 28.5 & 25.6 & 13.6$\quad$ & 23.7 & 13.5 & 22.8 & 29.3 & 24.7 & -$\quad$\\
	   & \cmark & {\textbf{27.7}} & {\textbf{21.8}} & {\textbf{28.0}} & {\textbf{29.7}} & {\textbf{28.2}} & {\textbf{8.8}}$\quad$ &  \textbf{26.5} & \textbf{18.5} & \textbf{26.5} & \textbf{30.2} & \textbf{28.3} & \textbf{9.0}$\quad$ \\
	   
	   \hline 
	   
	   \multirow{2}{*}{EQL v2 \cite{tan2021equalization}} & \xmark & {26.9} & {17.8} & {27.2} & {29.5} & {26.5} & 10.9$\quad$ & {25.5} & {17.7} & {24.3} & \textbf{30.2} & {26.1} & -$\quad$ \\
	   & \cmark & \textbf{27.8} & \textbf{20.9} & \textbf{28.4} & \textbf{29.9} & \textbf{28.1} & \textbf{7.4}$\quad$ &  \textbf{26.2} & \textbf{18.2} & \textbf{25.9} & {30.1} & \textbf{27.3} & \textbf{6.5}$\quad$\\
	   
	   \hline 
	   
	   \multirow{2}{*}{Seesaw \cite{wang2021seesaw}} & \xmark & {27.8} & {19.9} & {28.9} & {29.5} & {27.3} & 7.9$\quad$ & {26.8} & \textbf{19.8} & {26.3} & {30.5} & {27.6} & 7.5$\quad$ \\
	   
	   & \cmark & \textbf{28.8} & \textbf{23.4} & \textbf{29.6} & \textbf{30.0} & \textbf{28.6} & \textbf{7.0}$\quad$ & \textbf{27.2} & 19.0 & \textbf{27.1} & \textbf{30.9} & \textbf{28.1} & \textbf{6.4}$\quad$ \\
	   \hline 
   \end{tabular}
   \label{tab:benchmark_results}
\end{table*}

}
{
\noindent\textit{Extension to BCE classifier.} 
Beyond cross-entropy, binary cross-entropy (BCE) variants of classifiers with sigmoid activation function are also prevalent in modern instance segmentors. Our proposed method could be seamlessly applied without modification apart from discarding Equation \ref{eqa:cm_normalization} since the PCB regularization is designed for foreground classes, naturally matching the BCE context. Experiments on EQL v2 \cite{tan2021equalization} show the effectiveness.}

\section{Experiments}

In this section, we conduct experiments on the LVIS dataset \cite{gupta2019lvis} to validate the effectiveness of our method. We demonstrate the complementarity of our method to other state-of-the-art long-tailed instance segmentation methods.

\subsection{Experimental setup}

\noindent \textbf{Datasets.} Experiments are performed on the large vocabulary instance segmentation dataset LVIS. The LVIS v0.5 contains 1230 categories with both instance mask annotations and bounding box annotations. The latest version LVIS v1.0 includes 1203 categories. We conduct experiments and ablation studies mainly on LVIS v0.5, while the results on the prevalent LVIS v1.0 are also provided. The train set is used for training and the validation set is used for evaluation. All the categories are divided into three splits according to the number of images that each category appears in the train set: \textit{rare} (1-10 images), \textit{common} (11-100 images) and \textit{frequent} ($>$ 100).

\noindent \textbf{Evaluation metrics.} 
{Following common protocol}, we adopt Average Precision (AP) as the evaluation metric, which is averaged across \textit{IoU} threshold from 0.5 to 0.95. {AP for the main mask prediction task are omitted as AP} and AP for object detection is denoted as AP$^b$. AP50 and AP75 are also evaluated for comparison. We report detailed AP results on \textit{rare}, \textit{common} and \textit{frequent} splits, denoted as AP$^r$, AP$^c$ and AP$^f$. Besides, we also evaluate methods on a new metric to exam{ine} the pairwise class balance of confusion matrix on validation set, named Pairwise Bias (\textit{PwB} in short) . In detail, $PwB({M}) = \|{M} - {M}^\top \|_F$, where ${M}^\top$ stands for the transpose of ${M}$ and $\|\cdot\|_F$ is the Frobenius Normalization.

\begin{table*}
   \caption{Comparing with the state-of-the-art on LVIS v0.5 and LVIS v1.0 with various backbones. $^\dagger$ indicates results copied from~\cite{feng2021exploring}. $^\ddagger$ indicates results copied from \cite{chen2019mmdetection}.}
   \centering
   \small
   \begin{tabular}{lll|ccccccc}
      \hline
	    Dataset & Backbone & Method & AP & AP50 & AP75 & AP$_r$ & AP$_c$ & AP$_f$ & AP$^b$\\
	    \hline \hline
	    \multirow{5}{*}{LVIS v0.5} & \multirow{5}{*}{R-50-FPN} & BAGS~\cite{li2020overcoming}$^\dagger$ & 26.3 & - & - & 18.0 & 26.9 & 28.7 & 25.8 \\
	    & & EQL v2~\cite{tan2021equalization}$^\dagger$ & 26.9 & 41.5 & 28.9 &17.8 & 27.2 & 29.5 & 26.5\\
	    & & LOCE~\cite{feng2021exploring}$^\dagger$ & 28.4 & - & - &  22.0 & 29.0 & \textbf{30.2} & 28.2 \\

	   & & Seesaw~\cite{wang2021seesaw} & 27.8 & 42.6 & 29.6 & 19.9 & 28.9 & 29.5 & 27.3 \\
	    & & Seesaw + PCB & \textbf{28.8} & \textbf{43.8} & \textbf{30.9} & \textbf{23.4} & \textbf{29.6} & 30.0 & \textbf{28.6} \\
	   \hline 
	   \multirow{4}{*}{LVIS v1.0} & \multirow{4}{*}{R-50-FPN} & EQL v2~\cite{tan2021equalization}$^\dagger$ & 25.5 & - & - & 17.7 & 24.3 & 30.2 & 26.1 \\
	    & & LOCE~\cite{feng2021exploring}$^\dagger$ & 26.6 & - & - & 18.5 & 26.2 & 30.7 & 27.4 \\

	   & & Seesaw~\cite{wang2021seesaw}$^\ddagger$   & 26.8 & 41.3 & 28.4 & \textbf{19.8} & 26.3 & 30.5 & 27.6 \\
	    & & Seesaw + PCB & \textbf{27.2} & \textbf{41.7} & \textbf{29.4} & 19.0 & \textbf{27.1} & \textbf{30.9} & \textbf{28.1} \\
	    
	   \hline 
	   
	    \multirow{5}{*}{LVIS v1.0} & \multirow{5}{*}{R-101-FPN} &  BAGS~\cite{li2020overcoming}$^\dagger$ & 25.6 & - & - & 17.3 & 25.0 & 30.1 & 26.4 \\
	    & & EQL v2~\cite{tan2021equalization}$^\dagger$ & 27.2 & - & - & 20.6 & 25.9 & 31.4 & 27.9 \\
	    & & LOCE~\cite{feng2021exploring}$^\dagger$ & 28.0 & - & - & 19.5 & 27.8 & \textbf{32.0} & 29.0 \\

	   & & Seesaw~\cite{feng2021exploring}$^\ddagger$ & 28.2 & 42.7 & 30.2 & 21.0 & 27.8 & 31.8 & 28.9 \\
	    & & Seesaw + PCB & \textbf{28.8} & \textbf{43.3} & \textbf{30.9} & \textbf{22.6} & \textbf{28.3} & \textbf{32.0} & \textbf{29.9} \\
	   \hline 
   \end{tabular}
   \label{tab:comparison_with_sota}
\end{table*}

\noindent \textbf{Implementation details.}
For the implementation of our PCB, we set the dimension of hidden layers in $MLP_{loc}$ and $MLP_{cls}$ to 512. For EQL v2 \cite{tan2021equalization}, we addtionally apply a LayerNorm~\cite{ba2016layer} before $MLP_{cls}$ only to make the training stable, which will not increase the performance. For a RoI feature, it will go through the same prediction head for three iterations and generate classification scores and bounding box predictions three times, while only the last one is used for evaluation. The classification loss weights $w_t$ for each of the three iterations are set to 0.2, 0.2 and 0.6 as to guarantee the performance of the last iteration. The EMA momentum $\gamma$ is set to 0.99 as default, and we choose $\alpha=0.4$ for any method without modification of the loss function. For implementation on EQL v2 \cite{tan2021equalization} and Seesaw \cite{wang2021seesaw}, $L_{PCB}$ and $L_{CE}$ should be modified accordingly. As PCB regularization only replaces one-hot label to soft target, this could be easily achieved by applying this to these methods. Due to the change of loss function, the model is strongly re-balanced, $\alpha$ is to 0.2 and 0.05 respectively, as a complement. The PCB regularization term is  applied after the 16th epoch. For Seesaw~\cite{wang2021seesaw}, an RFS sampler is applied to have a fair comparison with LOCE \cite{feng2021exploring} which also resamples the data. For training strategies, we follow the common recipe~\cite{wang2021seesaw}, please refer to our supplemental material for details.

\subsection{Benchmark results}

Our PCB, which aims to achieve pairwise class balance, can be a complement to any existing long-tailed instance segmentation method. Experiments are conducted on LVIS v0.5 and LVIS v1.0 across various base methods, including two state-of-the-art solutions EQL v2 \cite{tan2021equalization} and Seesaw Loss \cite{wang2021seesaw}. Results are shown in Tab.~\ref{tab:benchmark_results}. When equipped with PCB, the AP of all base methods gets improved, especially for AP$_r$. Even on the strong re-balancing method Seesaw, there is still an obvious rise of the rare AP (e.g., + 3.5 AP$_r$ on LVIS v0.5). This is due to the exploration of the pairwise bias in PCB, which has not been considered in the previous works and thus still exists in those SOTA models. Another interesting thing is that PCB hardly ever hurts AP$_f$ to compensate for the AP$_r$, and the AP$_f$ is even improved sometimes. Additionally, the \textit{PwB} metric value and the model AP are negatively correlated, our PCB can effectively decreases the \textit{PwB}.

In Tab.~\ref{tab:comparison_with_sota}, we report the comparison with those state-of-the-art methods across different backbone networks on LVIS v0.5 and LVIS v1.0. In all experiments, the implemented Seesaw + PCB achieve both the best mask AP and box AP. The gap between $AP_r$ and $AP_f$ is narrowed.

\subsection{Ablation study}

In this part, A comprehensive ablation study is conducted to analyze various components in our PCB. Experiments are conducted on LVIS v0.5 using Mask R-CNN with ResNet-50-FPN. RFS is applied unless otherwise specified.

\noindent \textbf{Components in PCB.} There are two components in PCB, the regularization term and the learning paradigm. Experiments are conducted to verify the choice of each, results of which is summarized in Tab.~\ref{tab:component_ablation}. To evaluate the effectiveness of designed PCB regularization, we conduct a comparison with label smoothing~\cite{muller2019does} and online label smoothing~\cite{zhang2021delving}. Apparently, PCB regularization surpasses both. It obtains improvements on all splits, especially for the rare classes (+2.9 AP$_r$), which differs from traditional schemes doing trade-offs. On the contrary, online label smoothing which does not consider the long-tailed distribution exacerbates the bias. The comparison between deep supervision (DSN)~\cite{lee2015deeply, szegedy2015going} and our proposed iterative learning paradigm is also conducted. The performance of PCB gets improved when equipped with either DSN or iterative learning paradigm even if applying DSN alone hurts the performance, which shows the effectiveness of such a progressive learning manner. The prediction-based self-calibrated iterative learning paradigm still has advantages over DSN, the performance of all splits 
is better than that of which with DSN. The results of post-hoc confusion matrix calibration (CM) are also provided, of which the confusion matrix is calculated on the train set. The performance of PCB regularization surpasses the post-hoc confusion matrix calibration on all category splits, which supports our choice to do online learning.

\begin{table}
   \caption{Ablation study of each component in PCB. The choose of regularization (Regu.), and the learning paradigm (Paradigm). Label smoothing (LS)~\cite{muller2019does} and online label smoothing (OLS)~\cite{zhang2021delving} is to compare with PCB regularization. Deep supervision (DSN)~\cite{lee2015deeply} is to compare with our iterative learning paradigm. The results of post-hoc confusion matrix calibration (CM) are also provided. $^\dagger$ indicates using train set confusion matrix, which differs from that in Tab.~\ref{tab:motivation_tab}.}
   \centering
   \small
   \begin{tabular}{ll|ccccc}
      \hline
	     Regu. & Paradigm & AP & AP$_r$ & AP$_c$ & AP$_f$ & AP$^b$ \\
	   \hline \hline
	    N/A & N/A & {25.6} & {16.0} & {26.4} & {28.5} & {25.6} \\
	   CM$^\dagger$ & N/A & {25.4} & {17.8} & {25.8} & {27.8} & {25.2}\\
	   LS~\cite{muller2019does} & N/A & 25.9 & 16.9 & 26.3 & \textbf{29.1} & 26.0 \\
	   OLS~\cite{zhang2021delving} & N/A & 25.6 & 15.4 & 26.4 & 28.6 & 25.7 \\
	   PCB & N/A & \textbf{26.7}  & \textbf{18.9} & \textbf{27.5} & {28.8} & \textbf{27.1}\\
	   \hline
	   N/A & DSN~\cite{lee2015deeply} & {24.8} & {15.4} & {25.2} & {27.9} & {24.2} \\
	   N/A & Iterative & \textbf{26.5} & \textbf{17.4} & \textbf{27.3} & \textbf{29.2} & \textbf{26.8} \\
	   \hline
	   PCB & DSN~\cite{lee2015deeply} & {26.9} & {21.1} & {27.0} & {29.1} & {28.0} \\
	   PCB & Iterative & {\textbf{27.7}} & {\textbf{21.8}} & {\textbf{28.0}} & {\textbf{29.7}} & {\textbf{28.2}} \\
	   \hline 
   \end{tabular}
   \label{tab:component_ablation}
\end{table}

\noindent \textbf{Hyper-parameters.}
We test the two main hyper-parameters in PCB, the momentum $\gamma$ for the update of confusion matrix, and the PCB regularization coefficient $\alpha$. The results are summarized in Tab.~\ref{tab:gamma_analysis} and Tab.~\ref{tab:alpha_analysis}. In Tab.~\ref{tab:gamma_analysis}, with the increase of $\gamma$, the performance of rare classes is gradually improved, which is opposite to common classes. Overall, however, the performance is robust to the choice of $\gamma$. We vary $\alpha$ from 0 to 1. With the increase of $\alpha$, AP$_r$ first gets improved quickly then drops slowly. A similar phenomenon happens on AP$_c$ and AP$_f$ too. When $\alpha=0$, the model gets no regularization to relieve the bias. And if $\alpha=1$, too strong regularization harms the basic classification. We thus choose an $\alpha=0.4$ to trade-off between pairwise balance and discrimination.

\begin{table}
   \caption{Analysis on the influence of different EMA momentum $\gamma$.}
   \centering
   \small
   \begin{tabular}{c|ccccc}
      \hline
	    $\gamma$ & AP & AP$_r$ & AP$_c$ & AP$_f$ & AP$^b$ \\
	   \hline \hline
	   0.9 & 27.6 & {19.3} & \textbf{28.8} & {29.3} & {28.3} \\
	   0.99 & {\textbf{27.7}} & {21.8} & {28.0} & {\textbf{29.7}} & {28.2} \\
	   0.999 & {27.5} & \textbf{22.1} & {27.6} & {29.5} & \textbf{28.4}  \\
	   \hline 
   \end{tabular}
   \label{tab:gamma_analysis}
\end{table}

\begin{table}
   \caption{Analysis on the influence of different PCB regularization coefficient $\alpha$. Experiments are conducted with RFS on LVIS v0.5.}
   \centering
   \small
   \begin{tabular}{c|ccccc}
      \hline
	    $\alpha$ & AP & AP$_r$ & AP$_c$ & AP$_f$ & AP$^b$ \\
	   \hline \hline
	   0.0 & {26.5} & {17.4} & {27.3} & {29.2} & {26.8} \\
	   0.2 & {27.4} & {20.3} & {27.8} & {29.6} & {28.1} \\
	   0.4 & {\textbf{27.7}} & {\textbf{21.8}} & {28.0} & {\textbf{29.7}} & {28.2} \\
	   0.6 & {\textbf{27.7}} & {21.3} & {\textbf{28.2}} & {29.5} & {\textbf{28.5}}\\
	   0.8 & {27.4} & {21.3} & {27.9} & {29.1} & {28.4}\\
	   1.0 & {26.6}& {20.9}& {26.9}& {28.6} & {27.6} \\
	   \hline 
   \end{tabular}
   \label{tab:alpha_analysis}
\end{table}

\noindent \textbf{Prediction in each recurrent step.}
PCB outputs predictions at each recurrent step which is embedded to the input of next recurrent step. In Tab.~\ref{tab:iter_id_analysis}, we evaluate the classification predictions of each step. The performance of rare classes is gradually raised through iterations, while the performance of frequent classes is affected little or even improved. It indicates the mechanism of our PCB that, from recurrent step to recurrent step, the bias of the rare towards the frequent is relieved. The progressive debiasing manner ensures the performance of frequent classes. Interestingly, even the performance of the first recurrent step that is trained only by CE loss surpasses the corresponding base method in Tab.~\ref{tab:benchmark_results} and predictions from the middle step may even be comparable with the final prediction. It might be brought by the weight sharing of different recurrent steps.

\begin{table}
   \caption{The performance of each recurrent step's prediction of a 3-step PCB. Experiments are conducted on LVIS v0.5.}
   \centering
   \small
   \begin{tabular}{lc|ccccc}
      \hline
	    Method & step id & AP & AP$_r$ & AP$_c$ & AP$_f$ & AP$^b$ \\
	   \hline \hline
	   \multirow{3}{*}{Softmax} & 1 & {22.9} & {8.3} & {22.6} & {29.1} & {23.0} \\
	   & 2 & {24.6} & {10.6} & {25.0} & {\textbf{29.6}} & {24.7}  \\
	   & 3 & {\textbf{25.1}} & {\textbf{12.6}} & {\textbf{25.5}} & {29.5} & {\textbf{25.2}}  \\
	   \hline 
	   \multirow{3}{*}{RFS \cite{gupta2019lvis}} & 1 & {26.8} & {20.7} & {27.1} & {29.0} & {27.0} \\
	   & 2 & {\textbf{27.8}} & {21.0} & {\textbf{28.4}} & {\textbf{29.7}} & {\textbf{28.4}}  \\
	   & 3 & {27.7} & {\textbf{21.8}} & {28.0} & {\textbf{29.7}} & {28.2}  \\
	   \hline
   \end{tabular}
   \label{tab:iter_id_analysis}
\end{table}

\noindent \textbf{Comparison with other complementary methods.}
Besides our PCB, the recent proposed NORCAL \cite{pan2021model} is also a complementary method which does post-hoc calibration according to the training distribution. To show the superiority of PCB, experiments are conducted on various methods. The results are summarized in Tab.~\ref{tab:norcal_comparison} . It's not surprising that NORCAL can improve the performance of RFS while fails to do so on strong baselines. NORCAL only relies on the label distribution that lacks the inter-class relation, so will get embarrassed when the baseline is strongly re-balanced. It's not the situation for PCB, that PCB can still relieve the pairwise bias to obtain further improvements even on strong strong EQL v2 and Seesaw as shown in Tab.~\ref{tab:norcal_comparison}.

\begin{table}
   \caption{Comparison with NORCAL \cite{pan2021model} which is also complementary to other methods. NORCAL is grid searched and reported the results from optimal hyper-parameters for each method.}
   \centering
   \small
   \begin{tabular}{l|ccccc}
      \hline
	    Method & AP & AP$_r$ & AP$_c$ & AP$_f$ & AP$^b$ \\
	   \hline \hline
	   RFS & {25.6} & {16.0} & {26.4} & {28.5} & {25.6} \\
	   RFS + NORCAL & {27.4} & {19.6} & \textbf{28.8} & {28.8} & {27.4} \\
	   RFS + PCB & {\textbf{27.7}} & {\textbf{21.8}} & {28.0} & {\textbf{29.7}} & {\textbf{28.2}} \\
	   \hline 
	   EQL v2 & {26.9} & {17.8} & {27.2} & {29.5} & {26.5} \\
	   EQL v2 + NORCAL & {26.8} & {20.5} & {27.0} & {29.2} & {26.4} \\
	   EQL v2 + PCB & \textbf{27.8} & \textbf{20.9} & \textbf{28.4} & \textbf{29.9} & \textbf{28.1} \\
	   \hline
	   Seesaw & {27.8} & {19.9} & {28.9} & {29.5} & {27.3} \\
	   Seesaw + NORCAL & {27.8} & {20.9} & {28.9} & {29.2} & {27.4}\\
	   Seesaw + PCB & \textbf{28.8} & \textbf{23.4} & \textbf{29.6} & \textbf{30.0} & \textbf{28.6} \\
	   \hline 
   \end{tabular}
   \label{tab:norcal_comparison}
\end{table}

\section{Conclusions and Limitations}
In this paper, we propose to utilize the pairwise biases existing in the confusion matrix statistics as strong and intuitive indicator to facilitate more balanced long-tailed instance segmentation. Such indicator possesses more fine-grained inter-class relationship details which help achieve much higher performance upper bound. In short, an online calibration method aiming at Pairwise Class Balance (PCB) is proposed to relieve the long-tailed instance segmentation by generating fightback soft targets in the form of a simple regularization. Towards more friendly regularization, an iterative learning paradigm is devised to progressively relieve the pairwise bias. Experimentally, our proposed PCB method improves the performance of various existing long-tailed instance segmentation methods, {establishing} a new state-of-the-art on the very challenging LVIS benchmark. 

\noindent \textit{LIMITATION}: There could be room for improving statistical manner of the confusion matrix which is a tradeoff between historical and coming-batch statistics. The former introduces lags in reflecting current model status, while the later is unrepresentative. We will ameliorate PCB in the future.

{\small
\bibliographystyle{ieee_fullname}
\bibliography{egbib}
}

\newpage

\appendix

\section{Details in implementation}

\noindent \textbf{Implentation of MS calibration.} We found the original MS calibration mentioned in the main paper did not work well. The deep reason is: If instances of rare class $i$ are always predicted as frequent class $j$ with very high confidence (i.e., $M_{i,j} \approx 1$ and $M_{i, i} \approx 0$), then plenty of instances will be miscalibrated into class $i$ ($s_{i} \approx 0$). To avoid this, we did two slight modifications to the original MS calibration in our experiments. Firstly, instead of using $s_{i} = M_{i, i}$, we adopted $s_{i} = \sum_{k = 1}^C M_{k, i}$ to soften the distribution. Secondly, if $s_{i}$ was still close to 0, we did not predict class $i$ in this case. As shown in Table~\ref{tab:ms_modification}, the modifications do help improve the performance of MS calibration on each split.

\begin{table}[h]
   \caption{The performance of MS calibration before and after applied two modifications. Experiments are conducted on LVIS v0.5 using Mask R-CNN with ResNet-50-FPN and RFS~\cite{gupta2019lvis} sampler.}
   \centering
   \small
   \begin{tabular}{cc|ccccc}
      \hline
        \multicolumn{2}{c|}{Modification} & \multirow{2}{*}{AP} & \multirow{2}{*}{AP$_r$} & \multirow{2}{*}{AP$_c$} & \multirow{2}{*}{AP$_f$} & \multirow{2}{*}{AP$^b$} \\
	    \#1 & \#2 \\
	   \hline \hline
	    & \cmark & 20.0 & 13.5 & 19.1 & 23.7 & 19.9 \\
	   \cmark &  & 26.5 & 20.0 & 27.7 & 27.5 & 26.3 \\
	   \cmark & \cmark & \textbf{27.0} & \textbf{20.6} & \textbf{28.2} & \textbf{28.0} & \textbf{27.0} \\
	   \hline 
   \end{tabular}
   \label{tab:ms_modification}
\end{table}

\noindent \textbf{Applied to EQL v2 \cite{tan2021equalization} and Seesaw \cite{wang2021seesaw}.}
PCB regularization can be easily applied to EQL v2~\cite{tan2021equalization} and Seesaw loss~\cite{wang2021seesaw} as they only change the loss weight or model activation. For each $k$-th proposal of label $y$ at current iteration, the original equalization loss v2 can be formulated as follows:

\begin{equation}
    L_{EQL\,v2}(k) = - \sum_{i=1}^C w_i \left[y_i \log \hat{p}_i + (1 - y_i) \log (1 - \hat{p}_i)\right] \,,
\end{equation}
where $\hat{p}_i = 1/\left(1 + e^{-z_i^{fg}}\right)$ is the post-sigmoid probability for class $i$, $y_i$ indicates whether the proposal belongs to class $i$, and $w_i$ is a weight calculated from gradient perspective. When applied to EQL v2, our PCB regularization becomes
\begin{equation}
    L_{PCB} = - \sum_{i=1}^C w_i \left[\hat{M}_{i,y}^t \log \hat{p}_i + (1 - \hat{M}_{i,y}^t) \log (1 - \hat{p}_i)\right] \,.
\end{equation}
And the total classification loss function for the proposal is
\begin{equation}
    L_{cls}(k) = \alpha L_{PCB}(k) + (1 - \alpha)L_{EQL\,v2}\,.
\end{equation}
Similarly, we define the Seesaw variant of PCB regularization. The original Seesaw loss terms as 
\begin{align}
    L_{seesaw}(k) = - \sum_{i=1}^C y_i \log \hat{p}_i \,, \notag \\
    with\, \hat{p}_i = \frac{\exp(z_i^{fg})}{\sum_{j \neq i}^C \mathcal{S}_{ij} \exp(z_j^{fg}) + \exp(z_i^{fg})}\,. 
\end{align}
The $\mathcal{S}_{ij}$ is composed by a mitigation factor and a compensation factor. So, the PCB regularization can be written as 
\begin{align}
    L_{PCB}(k) = - \sum_{i=1}^C \hat{M}^t_{i, y} \log \hat{p}_i \,, \notag \\
    with\, \hat{p}_i = \frac{\exp(z_i^{fg})}{\sum_{j \neq i}^C \mathcal{S}_{ij} \exp(z_j^{fg}) + \exp(z_i^{fg})}\,. 
\end{align}
We combine them to get 
\begin{equation}
    L_{cls}(k) = \alpha L_{PCB}(k) + (1 - \alpha) L_{seesaw}\,.
\end{equation}

\noindent \textbf{Regression loss calculation.}
In practice, we calculate the regression loss only in the last recurrent step, rather than in each step. So the overall objective becomes:
\begin{equation}
    L = \left[\sum_{r=1}^R w_r L_{cls}^r\right] + L_{loc}^R + L_{mask}\,.
\end{equation}
It shows similar performance to the later, while obtains higher AP$_r$. The results of comparison are shown in Tab.~\ref{tab:analysis_on_regression}. While the overall mask AP and box AP are comparable for the two manners, there is a constant improvement in performance of rare classes (over 1 AP$_r$) for the former. 

\begin{table}
   \caption{Analysis on the influence of calculating the regression loss in each step or only last step. Experiments are conducted on LVIS v0.5.}
   \centering
   \small
   \begin{tabular}{cc|ccccc}
      \hline
	    Method & Regression & AP & AP$_r$ & AP$_c$ & AP$_f$ & AP$^b$ \\
	   \hline \hline
	   \multirow{2}{*}{Softmax} & each & \textbf{25.1} & 11.1 & \textbf{25.9} & \textbf{29.6} & \textbf{25.3} \\
	   & last & \textbf{25.1} & \textbf{12.6} & 25.5 & 29.5 & 25.2 \\
	   \hline
	   \multirow{2}{*}{RFS} & each & 27.5 & 20.4 & \textbf{28.2} & 29.4 & 27.8 \\
	   & last & \textbf{27.7} & \textbf{21.8} & 28.0 & \textbf{29.7} & \textbf{28.2} \\
	   \hline 
   \end{tabular}
   \label{tab:analysis_on_regression}
\end{table}

\noindent \textbf{Training details.} Following \cite{wang2021seesaw}, we implement our method with mmdetection \cite{chen2019mmdetection}. Mask R-CNN \cite{he2017mask} with ResNet-50-FPN and ResNet-101-FPN~\cite{he2016deep, lin2017feature} is adopted as our baseline model. We utilize the standard 2$\times$ schedule for LVIS of both versions. The models are trained using SGD with 0.9 momentum and 0.0001 weight decay for 24 epochs. With batch size of 16 on 8 GPUs, the initial learning rate is set to 0.02 and is decreased by 0.1 after 16 and 22 epochs, respectively. The training data augmentations include scale jitter{ing} (640-800) and horizontal flipping. For evaluation, we set the maximum number of detections per image to 300 and the minimum score threshold to 0.0001, as \cite{gupta2019lvis}.

\section{Analysis on $\alpha$ w/o iterative learning paradigm}
As discussed in Sec. 3.3 of the main paper, the performance will deteriorate soon with the increase of $\alpha$ if the iterative learning paradigm is not applied. Tab.~\ref{tab:alpha_analysis_pcb_regularization} shows an example. As $\alpha$ increases, AP$_r$ gets improved until $\alpha=0.6$, while AP$_c$ and AP$_f$ decline soon after $\alpha > 0.2$. So the PCB regularization hurts the fundamental classification, and the flexibility of debiasing is limited. By applying iterative learning paradigm, which guarantees the fundamental classification, such worry gets relieved. The room for debiasing is increased.

\begin{table}
   \caption{Analysis on the influence of different PCB regularization coefficient $\alpha$ without iterative learning paradigm. Experiments are conducted with RFS on LVIS v0.5.}
   \centering
   \small
   \begin{tabular}{c|ccccc}
      \hline
	    $\alpha$ & AP & AP$_r$ & AP$_c$ & AP$_f$ & AP$^b$ \\
	   \hline \hline
	   0.0 & 25.6 & 16.0 & 26.4 & 28.5 & 25.6\\
	   0.2 & \textbf{26.7} & 18.9 & \textbf{27.5} & \textbf{28.8} & \textbf{27.1}\\
	   0.4 & 26.5 & 18.3 & 27.4 & 28.7 & 26.7\\
	   0.6 & 26.2 & \textbf{19.7} & 26.4 & 28.4 & 26.6\\
	   0.8 & 26.0 & 18.5 & 26.7 & 28.1 & 26.5\\
	   1.0 & 25.1 & 18.7 & 25.9 & 26.5 & 25.3\\
	   \hline 
   \end{tabular}
   \label{tab:alpha_analysis_pcb_regularization}
\end{table}

\section{Comparing with refinement module in CrowdDet~\cite{chu2020detection}}
We notice that the refinement module (RM) in \cite{chu2020detection} is similar to our iterative learning paradigm. There are two main differences, RM concatenates the predictions and features rather than element-wise operation, and it utilizes features of the penultimate layer. We also provide the results of adopting RM as the learning paradigm, which are summarized in Tab.~\ref{tab:comparison_with_rm}. While RM achieves promising AP$_r$ compared to PCB regularization, it hurts the performance of common classes and frequent classes much, so the overall AP drops. 
Different from RM, our proposed iterative learning paradigm guarantees the performance of common and frequent classes.

\begin{table}
   \caption{Comparison with refine module (RM in short) in CrowdDet~\cite{chu2020detection}. Experiments are conducted with RFS on LVIS v0.5. PCB regularization is applied.}
   \centering
   \small
   \begin{tabular}{c|ccccc}
      \hline
	    Paradigm & AP & AP$_r$ & AP$_c$ & AP$_f$ & AP$^b$ \\
	   \hline \hline
	   N/A & 26.7 & 18.9 & 27.5 & 28.8 & 27.1 \\
	   RM & 26.4 & 20.8 & 26.5 & 28.5 & 26.9 \\
	   Iterative & \textbf{27.7} & \textbf{21.8} & \textbf{28.0} & \textbf{29.7} & \textbf{28.2} \\
	   \hline
   \end{tabular}
   \label{tab:comparison_with_rm}
\end{table}

\section{Extension to long-tailed classification.}
We also extend our PCB to long-tailed classification to testify its generalization ability. The commonly used ImageNet-LT~\cite{liu2019large} dataset is adopted in our experiment, and we use ResNeXt-50~\cite{xie2017aggregated} as the backbone network. Models are trained for 90 epochs with batch size 512. The intial learning rate is set to 0.2 and the first 5 epochs are trained with linear warm-up learning rate schedule~\cite{goyal2017accurate}. The learning rate is deacyed at 60$^{th}$ and 80$^{th}$ epoch by 0.1. For the implementation of PCB, we ignore the $MLP_{loc}$ and set the dimension of hidden layers in $MLP_{cls}$ to 256 for simplicity. For a feature vector from the backbone, it will go through the same classifier for two times, and the last prediction is used for evaluation. $\gamma$ is set to 0.9999. We train PCB in a decoupled manner~\cite{kang2019decoupling}, so the PCB regularizer is only applied in the fine-tune phase. 

Two methods are utilized as baseline, CE and BSCE~\cite{ren2020balanced}. The results are in Tab.~\ref{tab:extension_to_classification}. Equipped with PCB ($\alpha$ is set to 0.8 and 0.15 respectively), the performance gain is significant and consistent, the accuracy of few split is raised almost 10\% even on the strong baseline. The results fully demonstrate the generalization ability of our PCB.

\begin{table}
   \caption{Accuracy on ImageNet-LT with a ResNeXt-50 backbone.}
   \centering
   \small
   \begin{tabular}{cc|cccc}
      \hline
	    Method & PCB & Many & Medium & Few & Overall \\
	   \hline \hline
	   \multirow{2}{*}{CE} & \xmark & \textbf{67.76} & 38.89 & 7.44 & 45.73 \\
	   & \cmark & 61.68 & \textbf{49.10} & \textbf{21.97} & \textbf{50.26} \\
	   \hline
	   \multirow{2}{*}{BSCE~\cite{ren2020balanced}} & \xmark & \textbf{62.65} & 48.75 & 25.44 & 50.94 \\
	   & \cmark & 61.72 & \textbf{49.66} & \textbf{34.85} & \textbf{52.29} \\
	   \hline
   \end{tabular}
   \label{tab:extension_to_classification}
\end{table}

\end{document}